\documentclass{article}
\usepackage{arxiv}
\usepackage{times}
\usepackage{url}            
\usepackage{amsmath}
\usepackage{amssymb}
\usepackage{amsfonts}       
\usepackage{graphicx}
\usepackage{multicol}
\usepackage[hidelinks]{hyperref}

\newcommand{\ie}{\textit{i}.\textit{e}. }
\newcommand{\eg}{\textit{e}.\textit{g}. }

\newcommand{\bm}{\mathbf}

\newcommand{\bW}{\bm{W}}
\newcommand{\bw}{\bm{w}}
\newcommand{\bV}{\bm{V}}

\newcommand{\bx}{\bm{x}}
\newcommand{\bX}{\bm{X}}
\newcommand{\bh}{\bm{h}}
\newcommand{\bz}{\bm{z}}
\newcommand{\by}{\bm{y}}

\newcommand{\bb}{\bm{b}}

\begin{document}
\title{Sequence Prediction using Spectral RNNs}


\author{Moritz Wolter \\
        Institute for Computer Science, University of Bonn\\
        Fraunhofer Center for Machine Learning and SCAI \\
        {\tt\small wolter@cs.uni-bonn.de}\\
        \And J\"urgen Gall \\
        Institute for Computer Science, University of Bonn\\
        {\tt\small gall@cs.uni-bonn.de}
        \And  Angela Yao \\
        National University of Singapore\\
        {\tt\small ayao@comp.nus.edu.sg}
}

\maketitle

\begin{abstract}
Fourier methods have a long and proven track record as an excellent
tool in data processing. As memory and computational constraints gain importance in embedded and mobile applications, we propose to combine Fourier methods and recurrent neural network architectures. The short-time Fourier transform allows us to efficiently process multiple samples at a time. Additionally, weight reductions trough low pass filtering is possible. We predict time series data drawn from the chaotic Mackey-Glass differential equation and real-world power load and motion capture data.
\footnote{Source code available at \url{https://github.com/v0lta/Spectral-RNN}}
\end{abstract}

\section{Introduction}
Deployment of machine learning methods in embedded and real-time scenarios leads to challenging memory and computational constraints. We propose to use the short time Fourier transform (STFT) to make sequential data forecasting using recurrent neural networks (RNNs) more efficient. The STFT windows the data and moves each window into the frequency domain. 
A subsequent recurrent network processes multi sample windows instead of single data points and therefore runs at a lower clock rate, which reduces the total number of RNN cell evaluations per time step. Additionally working in the frequency domain allows parameter reductions trough low-pass filtering. Combining both ideas reduces the computational footprint significantly. We hope this reduction will help deploying RNNs on VR-devices for human pose prediction or on embedded systems used for load forecasting and management.

We show in our work that it is possible to propagate gradients through the STFT \emph{and} its inverse.  This means that we can use time domain-based losses that match the (temporal) evaluation measures, but still apply the RNN in the frequency domain, which imparts several computation and learning benefits.
Firstly, representations can often be more compact and informative in the frequency domain. The Fourier basis clusters the most important information in the low-frequency coefficients. It conveniently allows us to integrate a low-pass filter to not only reduce the representation size but also remove undesirable noise effects that may otherwise corrupt the learning of the RNN.

One of the challenges of working with the Fourier transform is that it requires handling of complex numbers.  While not yet mainstream, complex-valued representations have been integrated into deep convolutional architectures~\cite{Trabelsi}, RNNs~\cite{Arjovsky,Wisdom,wolter2018Complex} and Hopfield like networks \cite{goto2019chaotic}. Complex-valued networks require some additional overhead for book-keeping between the real and imaginary components of the weight matrices and vector states, but tend to be more parameter efficient. We compare fully complex and real valued-concatenation formulations.

Naively applying the RNN on a frame-by-frame basis does not always lead to smooth and coherent predictions~\cite{mao2019learning}.  When working with short time Fourier transformed (STFT) windows of data, smoothness can be built-in by design through low-pass filtering before the inverse short time Fourier transformation (iSTFT). 
Furthermore, by reducing the RNN's effective clockrate, we extend the memory capacity (which usually depends on the number of recurrent steps taken) and achieve computational gains.

In areas such as speech recognition~\cite{chan2016listen} and audio processing~\cite{dieleman2014end}, using the STFT is a common pre-processing step, \eg ~as a part of deriving cepstral coefficients.
In these areas, however, the interest is primarily for classification based on the spectrum coefficients and the complex phase is discarded, so there is no need for the inverse transform and the recovery of a temporal signal as output. We advocate the use of the forwards and inverse STFT directly before and after recurrent memory cells and propagate gradients through both the forwards and inverse transform.  
In summary, 
\begin{itemize}
    \item we propose a novel RNN architecture for analyzing temporal sequences using the STFT and its inverse.
    \item We investigate the effect of real and complex valued recurrent cells.
    \item We demonstrate how our windowed formulation of sequence prediction in the spectral domain, based on the reduction in data dimensionality and rate in RNN recursion can, significantly improve efficiency and reduce overall training and inference time.
\end{itemize}

\section{Related Works}
\label{sec:related_work}
Prior to the popular growth of deep learning approaches, analyzing time series and sequence data in the spectral domain, \ie~via a Fourier transform was a standard approach ~\cite{pintelon2012system}.  
In machine learning, Fourier analysis is mostly associated with signal-processing heavy domains such as speech recognition~\cite{chan2016listen}, biological signal processing~\cite{Minami1999RealTime}, medical image~\cite{Virtue} and audio-processing~\cite{dieleman2014end}. 
The Fourier transform has already been noted in the past for improving the computational efficiency of convolutional neural networks (CNNs)~\cite{bengio2007scaling,pratt2017fcnn}.  In CNNs, the efficiency comes from the duality of convolution and multiplication in space and frequency.  Fast GPU implementations of convolution, \eg in the NVIDIA cuDNN library~\cite{nvidia} use Fourier transforms to speed up their processing. Such gains are especially relevant for convolutional neural networks in 2D~\cite{bengio2007scaling,pratt2017fcnn} and even more so in 3D~\cite{wang2017winograd}.  However, the improvement in computational efficiency for the STFT-RNN combination comes from reductions in data dimensionality and RNN clock rate recursion. The Fourier transform has been used in neural networks in various contexts~\cite{zuo2008fourier}.  Particularly notable are networks, which have Fourier-like units~\cite{Godfrey2018decomposition} that decompose time series into constituent bases. Fourier-based pooling has been explored for CNNs as an alternative to standard spatial max-pooling~\cite{DFTpooling}.

Various methods for adjusting the rate of RNN recurrency have been explored in the past.  One example~\cite{lin1996learning} introduces additional recurrent connections with time lags.  Others apply different clock rates to different parts of the hidden layer~\cite{alpay2016learning,koutnik2014clockwork}, or apply recursion at multiple (time) scales and or hierarchies~\cite{chung2016hierarchical,serban2017multiresolution}. All of these approaches require changes to the basic RNN architecture.  Our proposal, however, is a simple alternative which does not require adding any new structures or connections.
Among others \cite{martinez2017human} approached mocap prediction using RNNs, while \cite{mao2019learning} applied a cosine transform CNN combination. In this paper we explore the combination of the complex valued windowed STFT and RNNs on mackey-glass, power-load and mocap time series data.

\section{Short Time Fourier Transform}
\label{sec:STFT}
\subsection{Forwards STFT}
The Fourier transform maps a signal into the spectral or frequency domain by decomposing it into a linear combination of sinusoids.  In its regular form the transform requires infinite support of the time signal to estimate the frequency response. For many real-world applications, however, including those which we wish to model with RNNs, this requirement is not only infeasible, but it may also be necessary to obtain insights into changes in the frequency spectrum as function of time or signal index.  To do so, one can partition the signal into overlapping segments and approximate the Fourier transform of each segment separately. This is the core idea behind the short time Fourier transform (STFT), it determines a signal's frequency domain representation as it changes over time.

More formally, given a signal $\bx$, we can partition it into segments of length $T$, extracted every $S$ time steps. The STFT $\mathcal{F}_s$ of $\bx$ is defined by~\cite{Griffin1984Estimation} as the discrete Fourier transform of $\bx$, \ie
\begin{align}
\begin{split}
    \bX [\omega, Sm]  &= \mathcal{F}_s\left(\bx\right) \\ 
    = \mathcal{F}\left(\bw[Sm - l]\bx[l]\right)
    &= \sum_{l = -\infty}^{\infty} \bw[Sm - l]\bx[l]e^{-j\omega l}
    \label{eq:STFT}
    \end{split}
\end{align}
\noindent where $\mathcal{F}$ denotes the classic discrete fast Fourier transform. Segments of $\bx$ are multiplied with the windowing function $\bw$ and transformed afterwards. 

Historically, the shape and width of the window function has been selected by hand. To hand the task of window selection to the optimizer, we work with a truncated Gaussian window~\cite{Harris1978windows}
\begin{align}
    w[n] = \exp\Big(-\frac{1}{2}\Big(\frac{n - T/2}{\sigma T/2}\Big)^2\Big)
\end{align}
\noindent of size T and learn the standard deviation $\sigma$. The larger that $\sigma$ is, the more the window function approaches a rectangular window; the smaller the sigma, the more extreme the edge tapering and as a result, the narrower the window width. 

\subsection{Inverse STFT}
Supposing that we are given some frequency signal $\bm{X}$;  
the time signal $\hat{\bx}$ represented by $\bm{X}$ can be recovered with the inverse short time Fourier transform (iSTFT) $\mathcal{F}_s^{-1}$ and is defined by~\cite{Griffin1984Estimation} as:
\begin{align}
    \begin{split}
    \hat{\bx}[n] & = \mathcal{F}_s^{-1}(\bX[n, Sm])
    = \frac{\sum_{m = -\infty}^{\infty} \bw[Sm - n]\hat{\bx}_w[n, Sm]}
    {\sum_{m = -\infty}^{\infty} {\bw}^2[Sm - n]}
    \label{eq:iSTFT}
    \end{split}
\end{align}
where the signal $\hat{\bx}_w$ is the inverse Fourier transform of $\bX$: 
\begin{align}
\hat{\bx}_w = \frac{1}{T}\sum_{l = -\infty}^{\infty} \bX[l, Sm] e^{j\omega l}
\end{align}
and $l$ indexing the frequency dimension of $\bX_m$. Eq.~\ref{eq:iSTFT} reverses the effect of the windowing function, but implementations require careful treatment of the denominator to prevent division by near-zero values\footnote{We adopt the common strategy of adding a small tolerance $\epsilon=0.001$}. In Eq.~\ref{eq:iSTFT}, $Sm$ generally evaluates to an integer smaller than the window size $T$ and subsequent elements in the sum overlap, hence the alternative naming of it being an \emph{``overlap and add''} method~\cite{grochenig2013foundations}. 

\section{Complex Spectral Recurrent Neural Networks}
\subsection{Network Structure}\label{sec:fRNN}
We can now move RNN processing into the frequency domain by applying the STFT to the input signal $x$. If the output or projection of the final hidden vector is also a signal of the temporal domain,
we can also apply the iSTFT to recover the output $y$.
This can be summarized by the following set of RNN equations:
\begin{align}
    \bX_\tau & = \mathcal{F}(\{\bx_{S\tau-T/2}, \dots,  \bx_{S\tau+T/2}\}) \label{eq:basicRNN21} \\
    \bz_{\tau} &= \bW_c \bh_{\tau-1} + \bV_c \bX_{\tau} + \bb_c \label{eq:basicRNN22} \\
    \bh_{\tau} &= f_a(\bz_{\tau}) \label{eq:basicRNN23}\\
    \by_{\tau} &= \mathcal{F}^{-1}(\{\bW_{pc}\bh_0, \dots, \bW_{pc}\bh_\tau \})
    \label{eq:basicRNN24}
\end{align}
where $\tau=[0,n_s]$, \ie~from zero to the total number of segments $n_s$.  The output $y_\tau$ may be computed based on the available outputs $\{\bW_p\bh_0, \dots, \bW_p\bh_\tau \}$ at step $\tau$. 
Adding the STFT-iSTFT pair has two key implications. First of all, because $\bX_\tau\in \mathbb{C}^{n_f \times 1}$ is a complex signal, the hidden state as well as subsequent weight matrices all become complex, \ie~$\bh_\tau \in \mathbb{C}^{n_h \times 1}$, $\bW_c \in \mathbb{C}^{n_h \times n_h}$, $\bV_c \in \mathbb{C}^{n_h \times n_f}$, $\bb_c \in \mathbb{C}^{n_h \times 1}$ and $\bW_{pc} \in \mathbb{C}^{n_h \times n_f}$, where $n_h$ is the hidden size of the network as before and $n_f$ is the number of frequencies in the STFT.

The second implication to note is that the step index changes from frame $t$ to $\tau$, which means that the spectral RNN effectively covers $S$ time steps of the standard RNN per step.  This has significant bearing on the overall memory consumption as well as the computational cost, both of which influence the overall network training time. 
Considering only the multiplication of the state matrix $\bW_c$ and the state vector $\bh_\tau$, which is the most expensive operation, the basic RNN requires $N \cdot O(n_h^3)$ operations for $N$ total time steps.  When using the Fourier RNN with an FFT implementation of the STFT, one requires only 
\begin{equation}\label{eq:complexity}
N/S \cdot (O(T \log T) + O(n_h^3)),
\end{equation}

\noindent where the $T\log T$ term comes from the FFT operation. The architectural changes lead to larger input layers and fewer RNN iterations. $\bX$ is higher dimensional than $\bx$, but we save on overall computation if the step size is large enough which will make $N/S$ much smaller than $N$. We can generalise the approach described above into:
\begin{align}\label{eq:genfrnn}
    \bX_\tau & = \mathcal{F}(\{\bx_{S\tau-T/2}, \dots,  \bx_{S\tau+T/2}\}) \\
    \bh_t &= \text{RNN}_\mathbb{C}(\bX_\tau, \bh_{t-1}) \\
    \by_t &= \mathcal{F}^{-1}(\{\bW_{pc}\bh_0, \dots, \bW_{pc}\bh_\tau \}),
\end{align}
where instead of the basic formulation outlined above, more sophisticated complex-valued  RNN-architectures~\cite{Arjovsky,Wisdom,wolter2018Complex} represented by $\text{RNN}_{\mathbb{C}}$ may be substituted. 
We experiment with a complex-valued GRU, as proposed in~\cite{wolter2018Complex}. An alternative to a complex approach is to concatenate the real and imaginary components into one (real-valued) hidden vector. The experimental section compares both methods in Table~\ref{tab:mackey_cvr}.

\subsection{Loss Functions}
In standard sequence prediction, the loss function applied is an L2 mean squared error, applied in the time domain: 
\begin{align}
    \mathcal{L}_{mse}(\textbf{y}_t, \textbf{y}_{gt}) = \frac{1}{n_y}\sum^{n_y}_{l=0}(\textbf{y}_t - \textbf{y}_{gt})^2.
\end{align}
We experiment with a similar variation applied to the STFT coefficients applied in the frequency domain (see Table~\ref{tab:mackey_cvr}) but find that it performs on par but usually a little bit worse than the time-domain MSE. This is not surprising, as the evaluation measure applied is still in the time domain for sequence prediction so it works best to use the same function as the loss.

\section{Mackey-Glass Chaotic Sequence Prediction}
\label{sec:mac_experiments}
\begin{figure}
    \centering
    \includegraphics[width=0.4\linewidth]{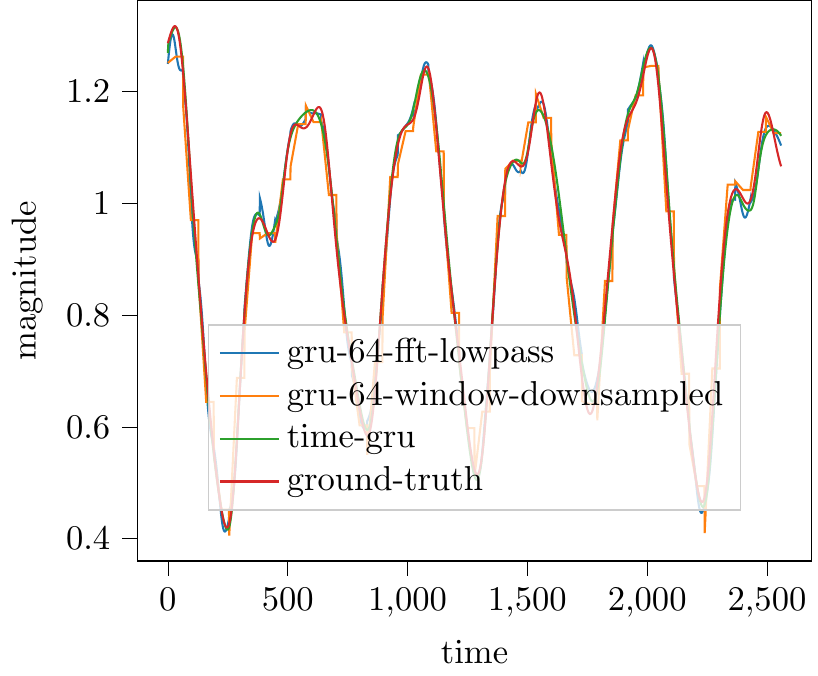}
    \includegraphics[width=0.4\linewidth]{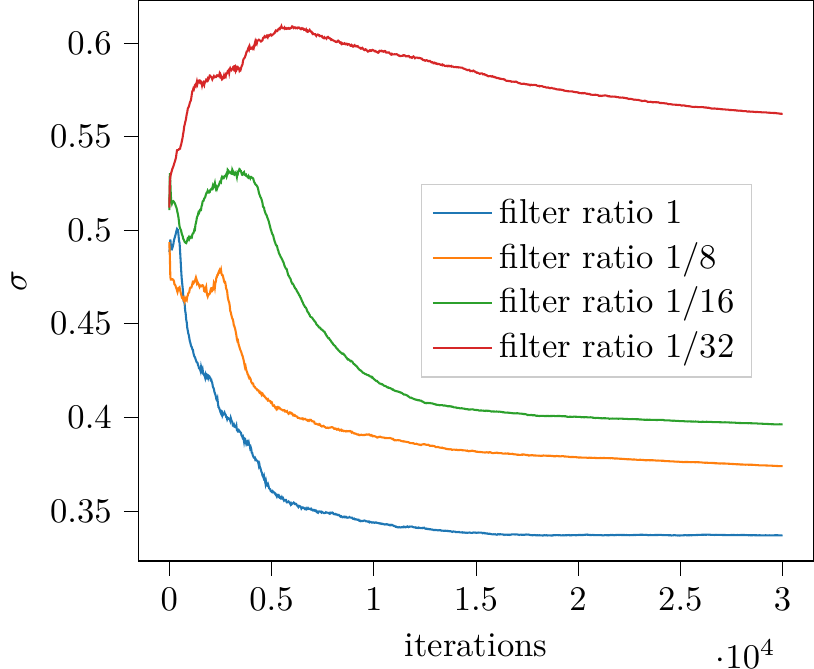}
    \caption{Mackey-Glass series predictions in for different RNN methods. As gradients flow through the STFT we can optimize the width of the gaussian $\sigma$. The learned window width for increasing degrees of low-pass filtering is shown here. Figure best viewed in colour.}
    \label{fig:mackey_results} 
\end{figure}

Initially we study our method by applying it to make predictions on the Mackey-Glass series~\cite{gers2002applying}. The Mackey-Glass equation is a non-linear time-delay differential and defines a chaotic system that is semi-periodic:
We first evaluate different aspects of our model on the chaotic Mackey-Glass time series~\cite{gers2002applying}:
\begin{equation}
    \frac{dx}{dt} = \frac{\beta x_\tau}{1 + x_\tau^n} - \gamma x,
\end{equation}
with 
$\gamma\!=\!0.1, \beta\!=\!0.2$ and the power parameter to $n=10$. $x_\tau$ denotes the value from $\tau$ time steps ago; we use a delay of $\tau=17$ and simulate the equation in the interval $t \in [0,512]$, using a forward Euler scheme with a time step of size $0.1$. During the warm-up, when true values are not yet known, we use a uniform distribution and randomly draw values from $1 + U[-0.1, 0.1]$. An example of the time series can be found in Figure~\ref{fig:mackey_results}; we split the signal in half, conditioning on the first half as input and predicting the second half.

\subsection{Implementation Details}
In all experiments, we use RNN architectures based on  real~\cite{cho-al-emnlp14} and complex~\cite{wolter2018Complex} GRU-cells with a state size of 64 and a Gaussian window of width 128 initialized at $\sigma=0.5$ unless otherwise stated.  The learning rate was set intially to 0.001 and then reduced with a stair-wise exponential decay with a decay rate of 0.9 every 1000 steps. Training was stopped after 30k iterations. Our run-time measurements include ordinary differential equation simulation, RNN execution as well as back-propagation time.

\subsection{Experimental Results and Ablation Studies}
\begin{table}[t]
    \centering
    \caption{Short time Fourier, Windowed and Time Domain results obtained using GRU cells of size 64. Windowed experiments process multiple samples of data without computing the STFT. Additionally we compare low-pass filtering the spectrum and downsampling the time domain windows. All models where trained for 30k iterations. We downsample and lowpass-filter with a factor of 1/32.}
    \begin{tabular}{c c c c}
    net                       & weights     & mse                &  training-time [min] \\ \hline
    time-GRU                    &  13k        & $3.8\cdot 10^{-4}$ &  355      \\
    time-GRU-window             &  29k        & $6.9\cdot 10^{-4}$ &  53       \\
    time-GRU-window-down    &  13k        & $12\cdot 10^{-4}$ &  48       \\
    STFT-GRU                &  46k        & $3.5\cdot 10^{-4}$ &  57       \\
    STFT-GRU-lowpass        &  14k        & $2.7\cdot 10^{-4}$ &  56       
    \end{tabular}
    \label{tab:mackey_results}
\end{table}
\begin{table}[t]
    \centering
    \caption{Real and complex valued architecture comparison on the mackey-glass data, with increasing complex cell size. The complex architectures take longer to run but are more parameter efficient. The last row shows a complex RNN cell in STFT space without iSTFT backpropagation.}
    \begin{tabular}{c c c c}
    net          & weights    & mse                 &  training-time [min] \\ \hline
    STFT-GRU-64   & 46k        & $3.5\cdot 10^{-4}$  & 57 \\
    STFT-cgRNN-32 & 23k        & $2.1\cdot 10^{-4}$  & 63 \\
    STFT-cgRNN-54 & 46k        & $1.6\cdot 10^{-4}$  & 63 \\
    STFT-cgRNN-64 & 58k        & $1.1\cdot 10^{-4}$  & 64 \\ \hline
    STFT-cgRNN-64-$\mathcal{L}_\mathcal{C}$ &  58k        & $210\cdot 10^{-4}$ &  64 \\
    \end{tabular}
    \label{tab:mackey_cvr}
\end{table}

\textbf{Fourier transform ablation}; We first compare against two time-based networks: a standard GRU (time-GRU) and a windowed version (time-GRU-windowed) in which we reshape the input and process windows of data together instead of single scalars. This effectively sets the clock rate of the RNN to be per window rather than per time step. As comparison, we look at the STFT-GRU combination as described in Section~\ref{sec:fRNN} with a GRU-cell and a low-pass filtered version keeping only the first four coefficients (STFT-GRU-lowpass). 
Additionally we compare low-pass filtering to time windowed time domain downsampling (time-GRU-window-down). For all five networks, we use a fixed hidden-state size of 64. From Figure~\ref{fig:mackey_results}, we observe that all five architecture variants are able to predict the overall time series trajectory. Results in Table~\ref{tab:mackey_results} indicate that reducing the RNN clock rate through windowing and parameters through low pass filtering also improves prediction quality. In comparison to time domain downsampling, frequency-domain lowpass filtering allows us to reduce parameters more 
aggressively.

\textbf{Runtime}; As discussed in Equation~\ref{eq:complexity}, windowing reduces the computational load significantly. In Table~\ref{fig:mackey_results}, we see that the windowed experiments run much faster than the naive approach. The STFT networks are slightly slower, since the Fourier Transform adds a small amount of computational overhead.

\textbf{The window function}; Figure~\ref{fig:mackey_results} shows the Gaussian width for multiple filtering scenarios. We observe a gradient signal on the window function. For more aggressive filtering the optimizer chooses a wider kernel.

\textbf{Complex vs. real cell}; We explore the effect of a complex valued cell in Table~\ref{tab:mackey_cvr}. We apply the complex GRU proposed in \cite{wolter2018Complex}, and compare to a real valued GRU. We speculate that complex cells are better suited to process Fourier representations due to their complex nature. Our observations show that both cells solve the task, while the complex cell does so with fewer parameters at a higher computational cost. The increased parameter efficiency of the complex cell could indicate that complex arithmetic is better suited than real arithmetic four Frequency domain machine learning. However due to the increased run-time we proceed using the concatenation approach.

\textbf{Time vs. Frequency mse}; One may argue that propagating gradients through the iSTFT is unnecessary. We tested this setting and show a result in the bottom row of Table~\ref{tab:mackey_cvr}. In comparison to the accuracy of the otherwise equivalent complex network shown above the second horizontal line, computing the error on the Fourier coefficients performs significantly worse. We conclude that if our metric is in the time domain the loss needs to be there too. We therefore require gradient propagation through the STFT.

\section{Power Load Forecasting}
\label{sec:power_load}
\subsection{Data}
We apply our Fourier RNN to the problem of forecasting power consumption. We use the power load data of 36 EU countries from 2011 to 2019 as provided by the European Network of Transmission System Operators for Electricity\footnote{\url{https://transparency.entsoe.eu/}; 
for reproducibility, we will release our crawled version online upon paper acceptance.}. 
The data is partitioned into two groups; countries reporting with a 15 minute frequency are used for day-ahead predictions, while those with hourly reports are used for longer-term predictions. 
For testing we hold back the German Tennet load recordings from 2015, all of Belgium's recordings of 2016, Austria's load of 2017 and finally the consumption of the German Ampiron grid in 2018. 

\begin{figure}[t]
\centering
\includegraphics[width=0.45\linewidth]{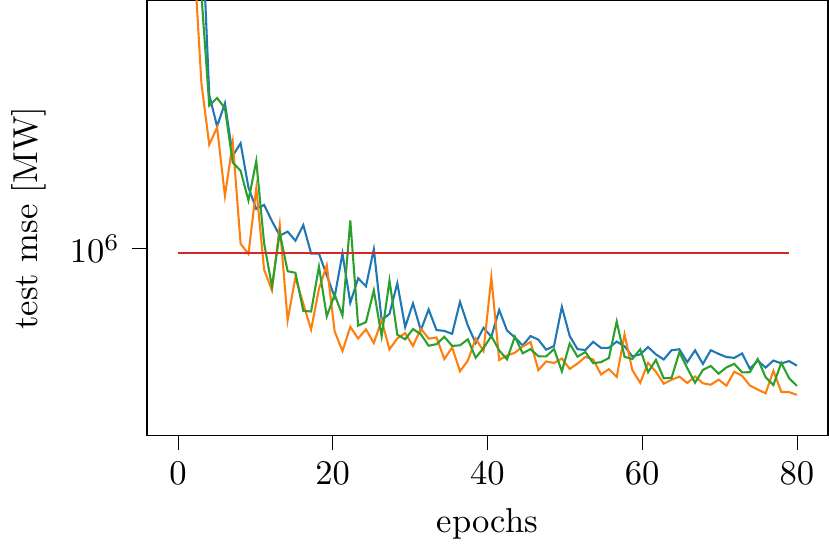}
\includegraphics[width=0.45\linewidth]{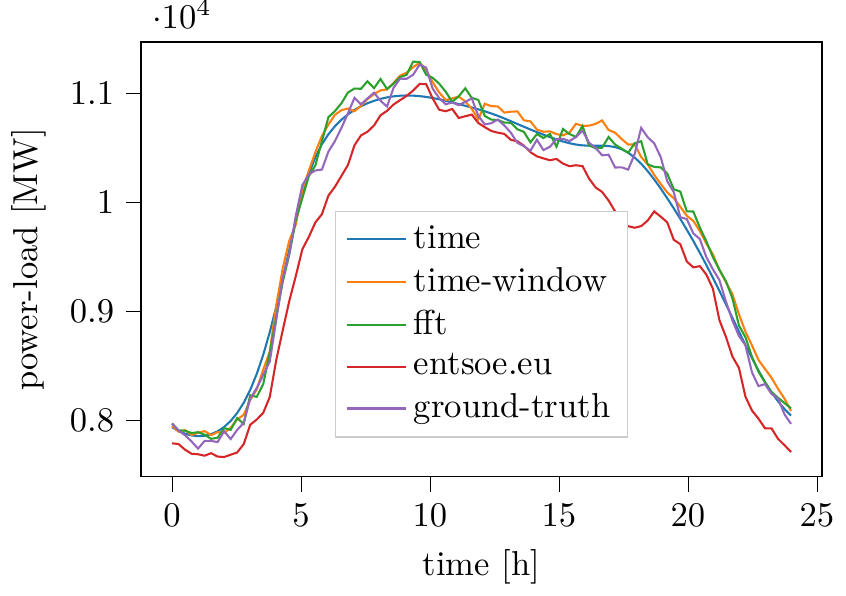}
\caption{Day ahead prediction results convergence (left) and prediction examples (right). We observe that all deep learning approaches beat the entsoe.eu baseline shown as the red line, which suggests that their approach could benefit from deep learning.}
\label{fig:res_day_ahead}
\end{figure}

\subsection{Day-Ahead Prediction}
We start with the task of day-ahead prediction; using 14 days of context, at 12:00, we are asked to predict 24 hours of load from midnight onwards (00:00 until 24:00 o' clock at 12:00 o'clock on the previous day). 
We therefore forecast the load from noon until midnight on the prediction day plus the next day and ignore the values from the prediction day. We use the same network architecture as in the previous section. During training, the initial learning rate was set to 0.004 and exponentially decayed using a decay of 0.95 every epoch. We train for 80 epochs overall. We compare time domain, windowed time as well as windowed Fourier approaches. The window size was set to 96 which corresponds to 24 hours of data at a sampling rate of 15 minutes per sample.  

In Figure~\ref{fig:res_day_ahead} we observe that all approaches we tried produce reasonable solutions and outperform the prediction produced by the European Network of Transmission System Operators for Electricity which suggests that their approach could benefit from deep learning.

\begin{table}[t]
\centering
\caption{60 day ahead power load prediction using GRUs of size 64. We downsample and lowpass-filter with a factor of 1/4. We observe that windowing leads to large training and inference speed-ups. Our STFT approach performs better in the full spectrum case and with a reduced input-dimensionality.}
\begin{tabular}{c c c c c c}
  Network                   & mse [MW]$^2$          & $\:$ weights $\:$ & $\:$ inference [ms] $\:$  & training-time [min] \\\hline 
  time-GRU                  & $261 \cdot 10^5$&  13k     & $1360$   & 1472 \\
  time-GRU-window           & $8.12 \cdot 10^5$&  74k    & $9.25$   & 15  \\
  time-GRU-window-down      & $8.05 \cdot 10^5$&  28k    & $8.2$    & 15  \\
  STFT-GRU                      & $7.62 \cdot 10^5$&  136k   & $9.67$   & 19  \\
  STFT-GRU-lowpass              & $7.25 \cdot 10^5$&  43k    & $9.69$   & 18  \\
\end{tabular}
\label{tab:60day_results}
\end{table}

\begin{figure}[t]
\centering
\includegraphics[width=0.45\linewidth]{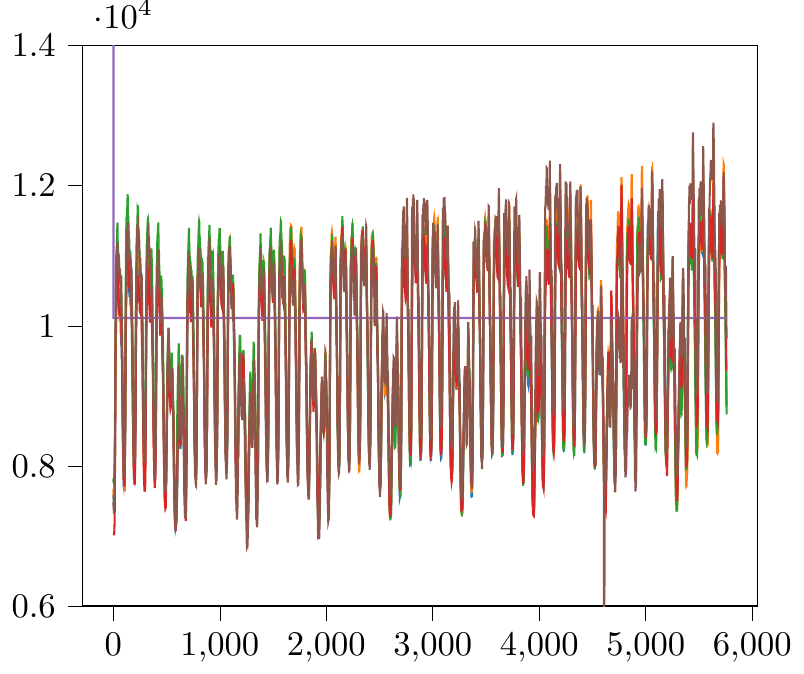}
\includegraphics[width=0.45\linewidth]{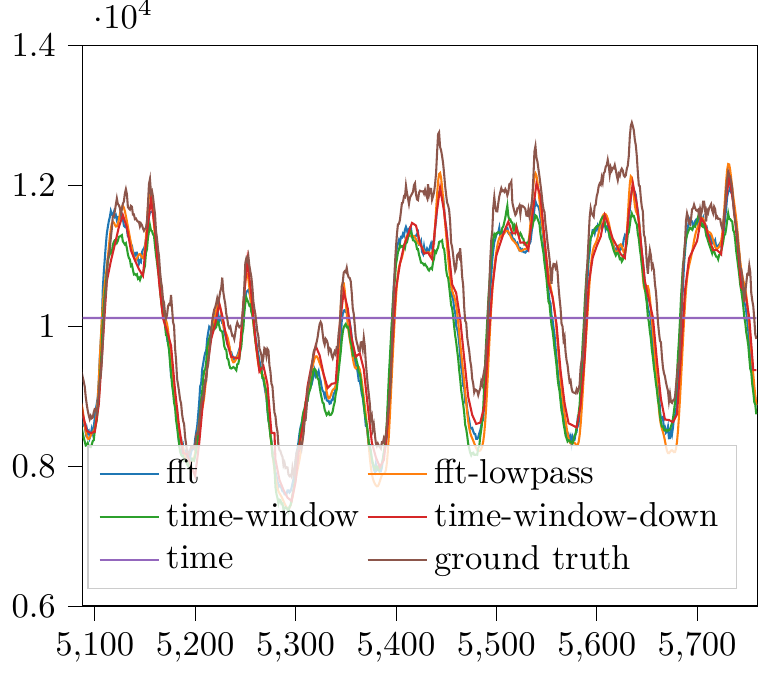}
\caption{A test set sample for showing the 60 day prediction results for all architectures under consideration. Close up for the last week of the 60 day prediction. }
\label{fig:60_day_fit}
\end{figure}

\subsection{Long-Term Forecast}
Next we consider the more challenging task of 60 day load prediction using 120 days of context. We use 12 load samples per day from all entenso-e member states from 2015 until 2019.
We choose a window size of 120 samples or five days; all other parameters are left the same as the day-ahead prediction setting. Table~\ref{tab:60day_results} shows that the windowed methods are able to extract patterns, but the scalar-time domain approach failed. Additionally we observe that we do not require the full set of Fourier coefficients to construct useful predictions on the power-data set. The results are tabulated in Table~\ref{tab:60day_results}. It turns out that the lower quarter of Fourier coefficients is enough, which allowed us to reduce the number of parameters considerably. Again we observe that Fourier low-pass filtering outperforms down-sampling and windowing.

\section{Human Motion Forecasting}
\subsection{Dataset \& Evaluation Measure}
The Human3.6M data set \cite{h36m_pami} contains 3.6 million human poses of seven human actors each performing 15 different tasks sampled at 50 Herz. As in previous work \cite{martinez2017human} we test on the data from actor 5 and train on all others. We work in Cartesian coordinates and predict the absolute 3D-position of 17 joints per frame, which means that we model the skeleton joint movements as well as global translation. 

\subsection{Implementation Details}
We use standard GRU-cells with a state size of 3072. We move the data into the frequency domain by computing a short time Fourier transform over each joint dimension separately using a window size of 24. The learning rate was set initially to 0.001 which we reduced using an exponential stair wise decay every thousand iterations by a factor of 0.97. Training was stopped after 19k iterations.
 
\subsection{Motion Forecasting Results}
\begin{figure}[t]
\centering
\includegraphics[width=0.495\linewidth]{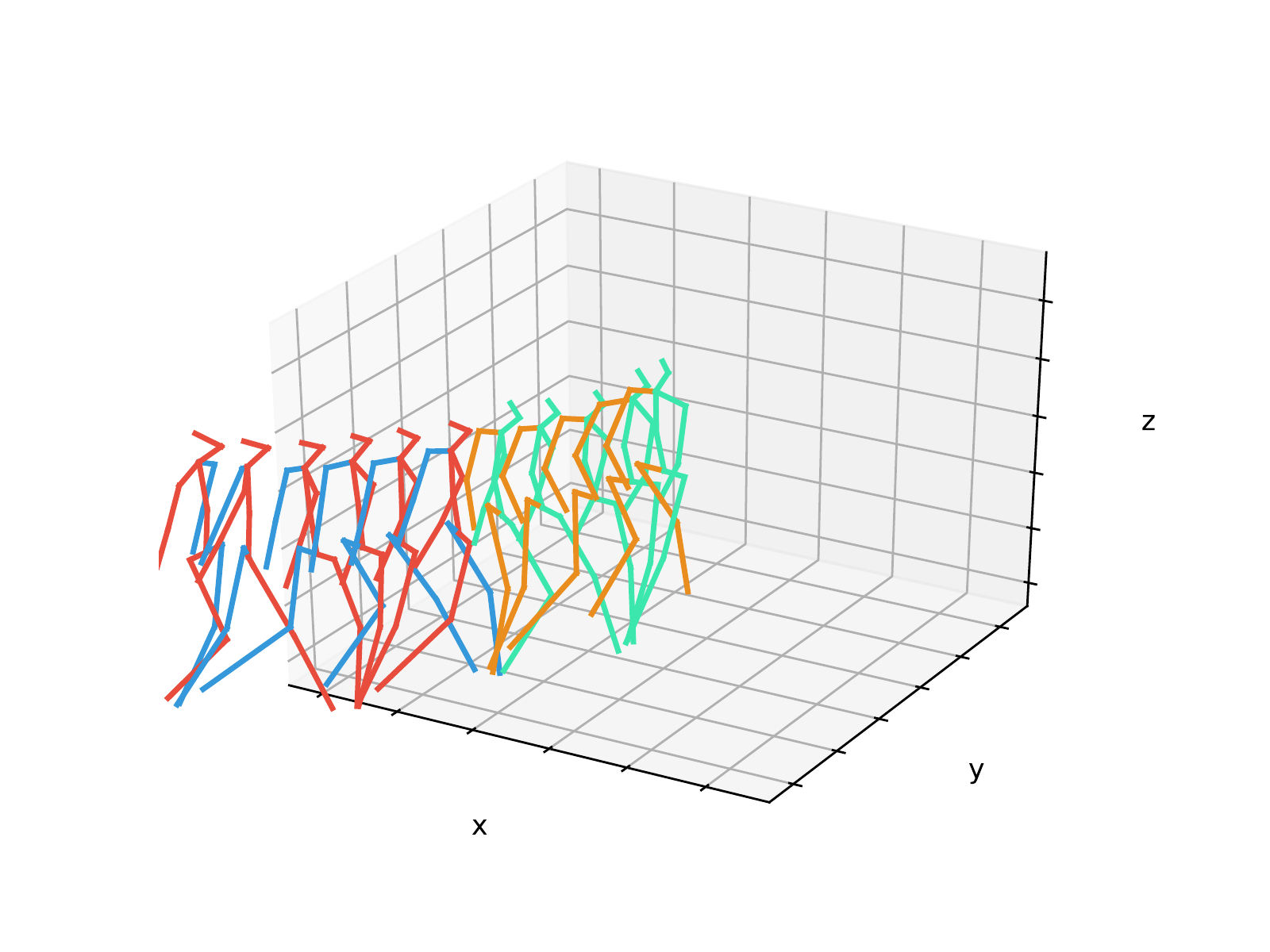}
\includegraphics[width=0.495\linewidth]{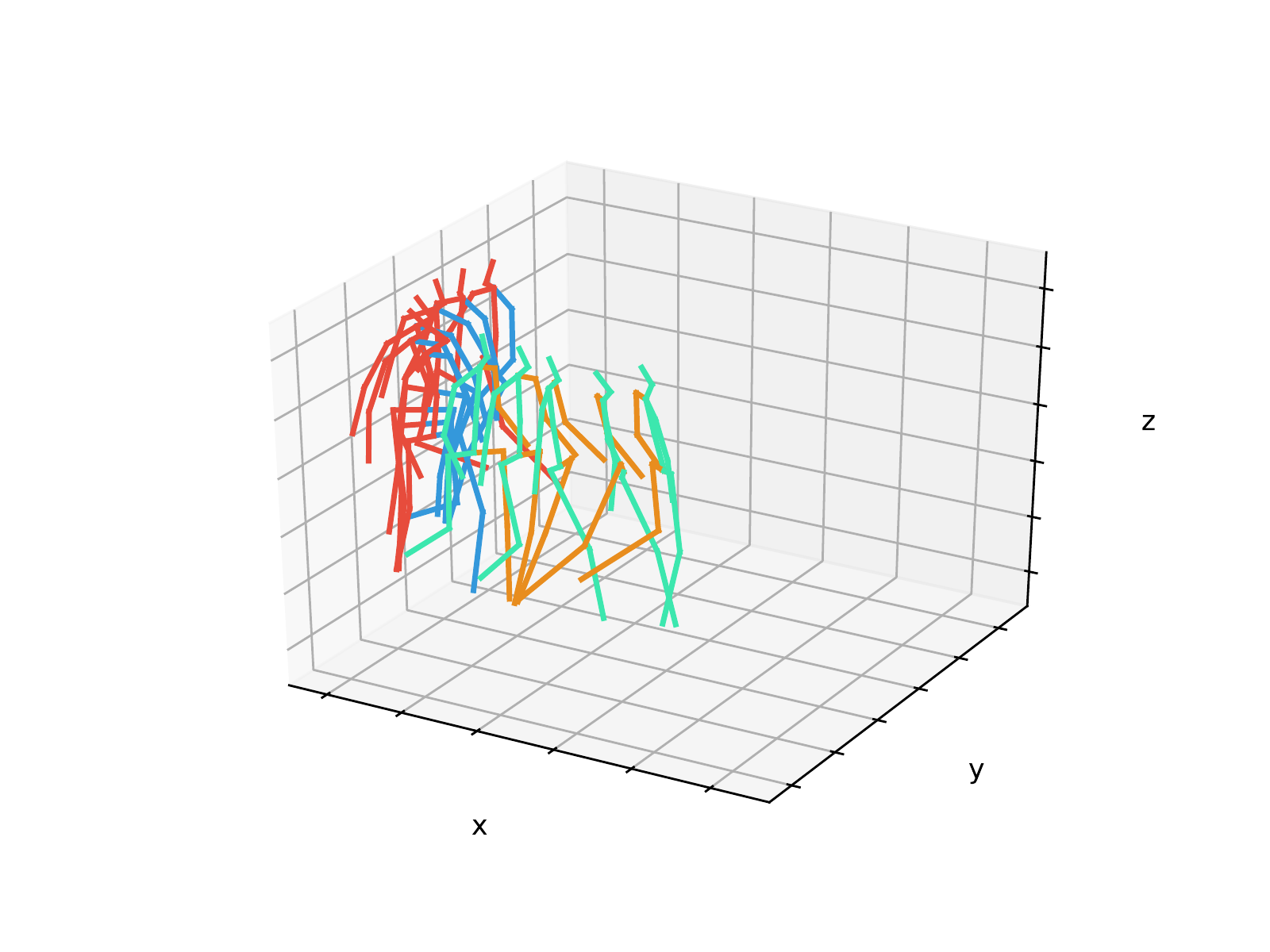}
\caption{Visualization of input and prediction results using a STFT-RNN combination and low pass filtering. Input is shown in red and blue, predictions in green and yellow.}
\label{fig:mocap_results}
\end{figure}
\begin{table}[t]
\caption{3d-Human motion forecast of 64 frames or approximately one second. Mean absolute error (mae) is measured in mm. Mean squared errors are reported in mm$^2$. We downsample and lowpass-filter with a factor of 1/4. Windowing runs much faster than the naive time domain approach. Among windowed approaches the STFT allows more aggressive input size reductions.}
\begin{tabular}{c c c c c c}
 Network   &$\:$mae [mm] $\:$&$\:$mse [mm]$^2$ $\:$&$\:$ weights $\:$&$\:$inference [ms]$\:$& training-time [min] \\\hline
 time-GRU \cite{martinez2017human}& 75.44 & $1.45\cdot 10^4$ &29M &	115         &   129           \\
 time-GRU-window     &	68.25     		 & $1.47 \cdot 10^4$ &33M &  18          &   27            \\
 time-GRU-window-down&	70.22     		 & $1.41 \cdot 10^4$ &30M &  18          &   27            \\
 STFT-GRU            &	67.88            & $1.25 \cdot 10^4$ &45M &  25          &   38            \\
 STFT-GRU-lowpass    &	66.71     		 & $1.30 \cdot 10^4$ &32M &  20          &   31            \\
\end{tabular}
\label{tab:mocap_results}
\end{table}
Prediction results using our STFT approach are shown in Figure~\ref{fig:mocap_results}. The poses drawn in green and yellow are predictions conditioned on the pose sequences set in blue and red. We observe that the predictions look realistic and smooth. In our experiments low-pass filtering helped us to enforce smoothness in the predictions.
Quantitative motion capture prediction results are shown in Table~\ref{tab:mocap_results}. We observe that all windowed approaches run approximately five times faster than the time domain approach during inference, a significant improvement. In terms of accuracy windowing does comparably well, while the STFT approach does better when the input sampling rate is reduced. 

\section{Conclusion}
In this paper we explored frequency domain machine learning using a recurrent neural network. We have proposed to integrate the Short Time Fourier transform and the inverse transform into RNN architectures and evaluated the performance of real and complex valued cells in this setting. We found that complex cells are more parameter efficient, but run slightly longer. Frequency domain RNNs allow us to learn window function parameters and make high-frequency time sequence predictions for both synthetic and real-world data while using less computation time and memory. Low-pass filtering reduced network parameters and outperformed time-domain down-sampling in our experiments.

\smallskip
\footnotesize{\noindent\textbf{Acknowledgements:} Work has been funded by the Deutsche Forschungsgemeinschaft (DFG, German Research Foundation) YA 447/2-1 and GA 1927/4-1 (FOR2535 Anticipating Human Behavior) as well as by the National Research Foundation of Singapore under its NRF Fellowship Programme [NRF-NRFFAI1-2019-0001].}

{\small
\bibliographystyle{unsrt}
\bibliography{fourier}
}
\end{document}